\pdfoutput=1

\documentclass[11pt]{article}

\usepackage[final]{acl}

\usepackage{amsmath,amsfonts,bm}

\def\eqref#1{equation~\ref{#1}}

\def\1{\bm{1}}

\DeclareMathAlphabet{\mathsfit}{\encodingdefault}{\sfdefault}{m}{sl}
\SetMathAlphabet{\mathsfit}{bold}{\encodingdefault}{\sfdefault}{bx}{n}

\usepackage{times}
\usepackage{latexsym}

\usepackage[T1]{fontenc}

\usepackage[utf8]{inputenc}

\usepackage{microtype}

\usepackage{inconsolata}

\usepackage{graphicx}

\usepackage{hyperref}
\usepackage{url}
\usepackage{lipsum}
\usepackage[ruled]{algorithm2e}
\usepackage{multicol}
\usepackage{latexsym}
\usepackage{booktabs}
\usepackage{wrapfig}
\usepackage{multirow}
\usepackage{caption}
\usepackage{enumitem}
\usepackage{tablefootnote}
\usepackage{array}
\usepackage{svg}
\usepackage{tcolorbox}
\usepackage{amsmath}
\usepackage{setspace}

\usepackage[noend]{algpseudocode}

\usepackage{xcolor}
\hypersetup{
    colorlinks,
    linkcolor={red!50!black},
    citecolor={blue!50!black},
    urlcolor={blue!80!black}
}
\usepackage[noabbrev,capitalize]{cleveref}

\crefname{equation}{equation}{equations}
\crefname{line}{line}{lines}
\crefname{section}{\S}{\S\S}
\Crefformat{section}{#2\S#1#3}
\crefrangeformat{section}{\S\S#3#1#4--#5#2#6}

\newcommand{\method}[1]{Velocitune}
\newcommand{\mathdata}{Reasoning}

\title{\method{}: A Velocity-based Dynamic Domain Reweighting Method for Continual Pre-training}

\author{
Zheheng Luo$^{\spadesuit\dagger}$\thanks{The first two authors contribute equally. Work done during an internship at Microsoft. Correspondence: \texttt{zhehengluo@gmail.com};~~\texttt{xinzhang3@microsoft.com}}~~~~
{\bf Xin Zhang$^{\dagger}$}\footnotemark[1]~~~~  
{\bf Xiao Liu}$^{\dagger}$\\ 
{\bf Haoling Li}$^{\diamondsuit\dagger}$~~~~ 
{\bf Yeyun Gong}$^{\dagger}$~~~~ 
{\bf Chen Qi}$^{\dagger}$~~~~ 
{\bf Peng Cheng}$^{\dagger}$ \\ 
$^\spadesuit$The University of Manchester\quad $^\dagger$Microsoft\quad
$^\diamond$Tsinghua University
}

\begin{document}
\maketitle

\begin{abstract}
It is well-known that a diverse corpus is critical for training large language models, which are typically constructed from a mixture of various domains.
In general, previous efforts resort to either sampling training data from different domains with static proportions or dynamically adjusting these proportions during training to optimise pretraining performance.
However, few methods addressed the complexity of domain-adaptive continual pre-training. 
To fill this gap, we propose \method{}, a novel framework that dynamically assesses learning velocity and adjusts data proportions accordingly, favouring slower learning domains while de-emphasising faster learning ones, which is guided by a scaling law to estimate the desired learning goal for each domain with a less associated cost.
To evaluate the effectiveness of \method{}, we conduct experiments on a dataset focused on reasoning tasks with CodeLlama, as well as on a corpus of system commands using Llama3 and Mistral.
\method{} achieves performance gains in both math and code reasoning tasks and command-line generation benchmarks.
Further analysis reveals that key factors driving the effectiveness of \method{} include target estimation and data ordering.
\end{abstract}

\section{Introduction}
\label{sec:intro}

Datasets used for pre-training language models (LMs) are typically composed of texts of various meta-attributes such as source and focus, referred to as different domains \citep{du2022glam,azerbayev2023llemma,together2023redpajama}.
The distinct characteristics of data from these varying domains, such as focus, quality, and quantity, affect the downstream performance of LMs differently \citep{rozière2024codellamaopenfoundation, li2024datacomplmsearchgenerationtraining}.
Consequently, numerous studies have explored the optimal combination of data from multiple domains to enhance LM performance.
\begin{figure*}[t]
    \centering
    \includegraphics[width=1\linewidth]{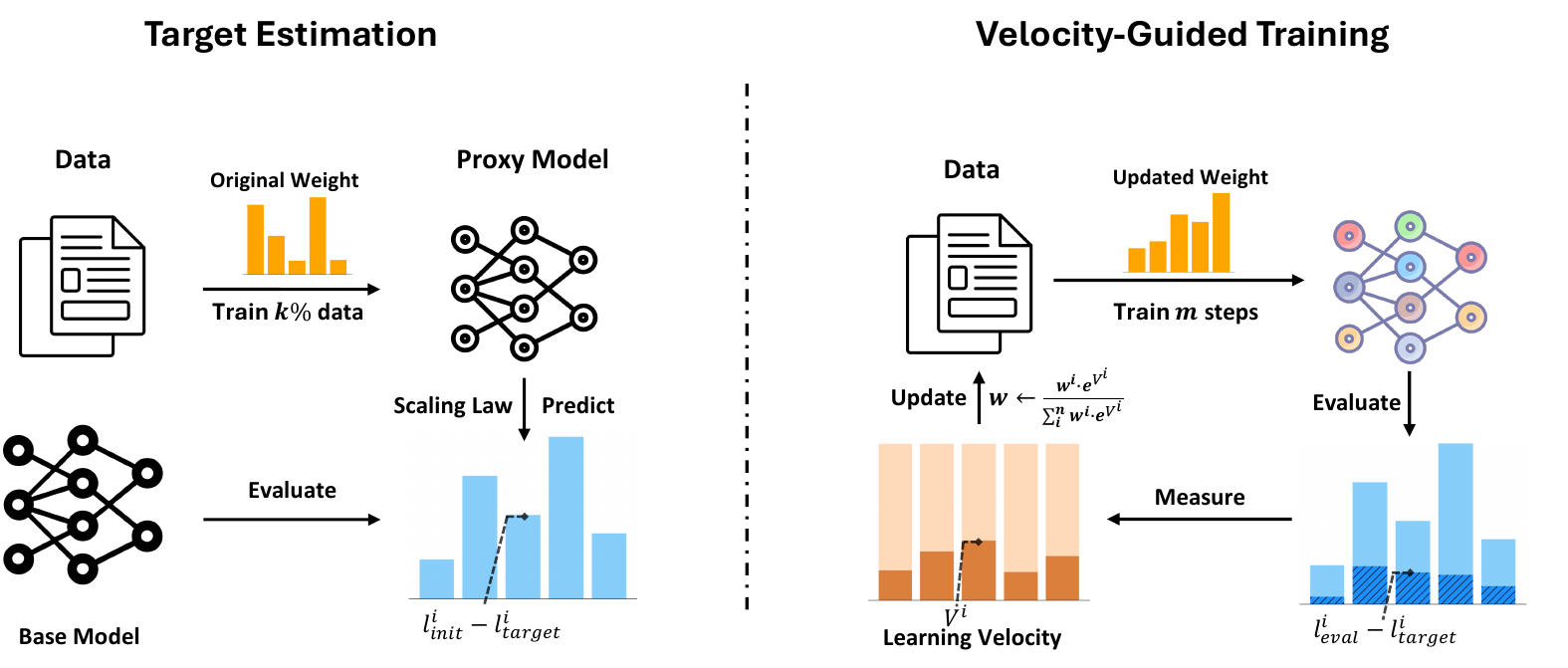}
    \caption{The overall pipeline of \method{}. Initially, a proxy model is trained using the original domain weights on a subset of the data. Following this, the initial loss is collected by evaluating the base model, while the target loss is determined by extrapolating the evaluation loss of the proxy model. In the second phase, we calculate the learning velocity by rescaling the learning progress between the initial and target losses. This learning velocity is then used to update the domain weights effectively.}
    \label{fig:pipeline}
\end{figure*}
Llama3~\citep{llama3modelcard}, GLaM~\citep{du2022glam}, and Lemma~\citep{azerbayev2023llemma} employ heuristic methods to iteratively test different ratios by training multiple proxy models and selecting the mixture that demonstrates the best downstream performance.
However, these heuristic approaches demand costly large-scale experiments for effective exploration. 
As a result, recent research is focused on learning an optimal ratio by dynamically adjusting weights during proxy model training~\citep{xie2024doremi, fan2023doge}.  For example,  Doremi~\citep{xie2024doremi} implements a method in which a small reference model is initially trained, followed by training a small proxy model using group distributionally robust optimisation (Group DRO)~\citep{Sagawa2019DistributionallyRN} to obtain optimised domain weights. 

Domain-adaptive\footnote{Here, "domain-adaptive" refers to optimising a model's ability and knowledge in a specific field or area. It should not be confused with the concept of "domain" of data as discussed in this paper.} continual pre-training, while sharing some similarities with from-scratch training, presents unique challenges that limit the effectiveness of existing domain reweighting methods.
Many existing methods utilise small proxy models to estimate optimal domain weights, which are subsequently transferred to a larger model \citep{xie2024doremi, fan2023doge, azerbayev2023llemma}. However, this approach poses challenges in domain-adaptive continual pre-training, as smaller versions of the base model often do not exist, making it difficult to estimate weights from proxy models.
Another challenge is how to leverage the learning status. Previous methods rely on the distance between the current loss and the target loss of the model for a domain ~\citep{xie2024doremi, xia2023sheared}. However, this distance-based approach can result in overemphasis on specific domains, as domains with larger loss disparities may disproportionately influence learning. This can exacerbate imbalances across domains.

To address these issues, we introduce a novel framework, \method{}, centred on the concept of learning velocity, as illustrated in \cref{fig:pipeline}. In contrast to previous approaches that leverage the distance between the current loss and the target loss, \method{} more effectively captures how fast models learn in each domain by establishing the learning velocity.
During training, domains exhibiting slower learning velocities are given increased weights, while those with faster velocities receive reduced weights, ensuring balanced learning progress.

To quantify learning velocity, it is crucial to determine both the model's already learnt expertise and its desired learning goal for each domain.
In the absence of a smaller proxy model, we leverage the Chinchilla scaling law~\cite{hoffmann2022training}, using the loss recorded on sub-sampled training data to cost-effectively predict the learning goal. 

We evaluated the performance of \method{} in two settings: continual pre-training CodeLlama 7B~\citep{rozière2024codellamaopenfoundation} on a math and coding reasoning dataset, as well as
Llama3 and Mistral~\citep{jiang2023mistral} on a system command knowledge corpus.
\method{} demonstrates an average improvement of 1.6\% in eight math tasks and 3.8\% in two coding tasks compared to the baseline trained with default weights. 
In addition, \method{} outperforms the baselines in Llama3, showing improvements of 4.9\% and 3.1\%, and in Mistral, with gains of 4.4\% and 2.6\% in the CmdGen-NVIDIA and CmdGen-AMD tasks, respectively. In addition, we conducted an in-depth ablation study to identify key factors contributing to the observed improvements. Our findings indicate that, beyond the contribution of the reweighted data ratios, the sequence of data ordering might also play a significant role in the effectiveness of \method{}. The results show that incorporating the predicted target loss is critical for the effectiveness of the learning velocity.
The contributions of this work can be summarised as follows. 
\begin{itemize}[leftmargin=*]
\item Introducing Velocitune, a novel framework for dynamically adjusting data ratios in continual pretraining. Velocitune estimates learning velocity to more precisely assess learning progress across domains and leverages scaling laws to optimise data allocation while minimising costs.

\item Demonstrating through extensive experiments that Velocitune enhances downstream performance in two continual pre-training settings. 
\item Providing a detailed analysis revealing that reweighted data ratios, predicted target loss, and data ordering contribute to the effectiveness of Velocitune.
\end{itemize}
\section{Domain-adaptive continual pre-training with \method{}}
\label{sec:method}
In this section, we present a detailed explanation of \method{}. Our approach focusses on adjusting domain weights based on learning velocity.
To quantify learning velocity, it is essential to determine the model’s existing expertise in each domain, which can be represented by its initial evaluation losses. Additionally, we define a learning target as the expected loss that the model should achieve given a certain amount of training data. This target provides a reference for measuring progress and adjusting domain weights accordingly.
During training, we periodically assess learning velocity, increasing the sampling weights of slower domains while reducing the weights of faster ones. This adjustment ensures balanced learning progress across domains. The methodology is detailed in \cref{algorithm}.
\subsection{The \method{} algorithm}
\label{algorithm}
\paragraph{Setup} Consider training a language model on a dataset $S$ consisting of $n$ distinct domains, denoted $D_1, D_2, \dots, D_n$. $ S_1, S_2, \dots, S_n$ represent the subsets of data corresponding to each domain, where each $S_i$ is divided into a train and evaluation set, denoted as $S_{\text{train}, i}$ and $S_{\text{eval}, i}$, respectively. The weight of the domain $w \in \Delta^n$ represents the sampling weight assigned to the domains.
\paragraph{Target Estimation} We first  evaluate the LM on each evaluation set $S_{\text{eval,},i}$ before training to obtain the initial loss $\ell_\text{init}$, providing an accurate measure of the model's learnt expertise in each domain. The target loss $\ell_\text{target}$ for each domain is derived using the scaling law~\citep{hoffmann2022training}. 
Specifically, we apply the Chinchilla scaling law by first defining the total dataset size and then training the model on subsets of training data using the default domain weights, which correspond to the ratio of tokens in each domain. Throughout the training process, multiple checkpoints are saved and the evaluation losses from these checkpoints are used to fit the scaling law parameters. Finally, we take the fitted function to predict the loss that the model could reach for using the entire dataset.  Details of the implementation and analysis of the prediction errors are provided in \cref{appd: scalaw}.



\paragraph{Velocity-Guided Training} After obtaining the initial loss and target loss in each domain, \method{} iteratively updates domain weights, denoted by \(w_t\), to minimise the re-weighted training loss effectively.
The optimisation problem is formulated as follows, in two sequential steps:

\noindent(1) Minimise the weighted training loss:
\begin{equation}
   \min_\theta \sum_{t=1}^T \sum_{i=1}^n w_t[i] \cdot \ell_t^i(\theta)
\end{equation}

where $T$ is the total training steps,  \(\ell_t^i(\theta)\) denotes the training loss at step \(t\) for domain \(D_i\) with model parameters \(\theta\). The \(w_t[i]\) refers to the weight assigned to domain \(D_i\) at step \(t\).

\noindent(2) Minimise the weighted sum of learning velocities:
\begin{equation}
\min_{w \in \Delta^n} \sum_{i=1}^n w_t[i] \cdot 
   V_t[i]
\end{equation}
$V_t[i]$ is the learning velocity of domain \(D_i\) at step \(t\).
The first step is to minimise the language modelling training loss given the domain weights. The second step focusses on dynamic adjustment so that the weight in the slowest learning velocity domain is maximised. 
Unlike Doremi~\citep{xie2024doremi}, which updates domain weights based on the loss difference of each data point between a proxy model and a trained reference model, our method does not rely on a fully trained reference model. Consequently, the loss of each data point in the trained model is unobtainable. Instead, after a set number of training steps, we update the domain weights by first calculating the learning velocity, which reflects how quickly the model is learning in a given domain. The learning velocity is defined as:

\begin{equation}
\label{eq:velocity}
V_t[i] = \frac{\ell_t^i(\theta, S^{\prime}_{\text{eval}, i}) - \ell_{\text{target}}^i}{\ell_{\text{init}}^i - \ell_{\text{target}}^i}
\end{equation}

Here $S^{\prime}_\text{eval}$ is a subset of $S_\text{eval}$ to speed up velocity estimation during training and $\ell_t^i(\theta, S^{\prime}_{\text{eval}, i})$ is the evaluation loss on $S^{\prime}_\text{eval}$ at step $t$ in domain \(D_i\). Then the domain weights are updated every $m$ steps exponentially following Group DRO \cite{sagawa2019distributionally}:

\begin{equation}
  \label{eq:updateW}
w_t \leftarrow \frac{w_{t-m}[i] \cdot \exp(V_t[i])}{\sum_{i=1}^n w_{t-m}[i] \cdot \exp(V_t[i])}
\end{equation}

Equation \ref{eq:updateW} adjusts the weights based on the learning velocity in each domain, ensuring an equitable comparison of the learning efficiency between domains with varying initial and target losses. By doing so, Velocitune prioritises learning in domains where progress towards the target loss is most promising, thereby enhancing overall model performance.
In summary, this approach dynamically rebalances the learning process across multiple domains, adjusting weights in response to the observed rate of learning progress, and steering them towards achieving optimal and uniform performance improvements. The complete \method{} algorithm is summarised in Algorithm \ref{alg:domrescal}.

\makeatletter
\newcommand{\removelatexerror}{\let\@latex@error\@gobble}
\makeatother

\begin{figure}[t]
  \centering
\removelatexerror
  \begin{algorithm}[H]
  \SetKwFunction{clamp}{Clamp}
  \SetKwFunction{update}{UpdateWeight}
  \SetKwProg{sub}{Function}{}{}
  \textbf{Require}: Data of $n$ domains $S_{1}, S_{2},\cdots,S_{ n}$, initial data loading weights $w_0 \in \mathbb{R}^n$ initialise as uniform distribution, initial loss $\ell_\mathrm{init} \in \mathbb{R}^n$, target loss $\ell_\mathrm{target} \in \mathbb{R}^n$, $\ell_t^i(\theta, S^{\prime}_{\text{eval},i})$ evaluation loss of the model at time $t$ , evaluation interval $m$, model parameters $\theta$. \\
\vspace{0.5em}
  \For{$t=1, \cdots, T$}{
    \If{$t \mod m = 0$} {
      $\Delta_t[i] \gets \clamp \left\{0 , \frac{(\ell_t^i(\theta, S^{\prime}_{\text{eval}, i}) - \ell_{\mathrm{target}}[i] )}{  (\ell_{\mathrm{init}}[i] - \ell_{\mathrm{target}}[i])}, 1 \right\}$ \Comment{Measure learning velocity} \\
      $w_t\gets$ \update{$w_{t-m}$, $\Delta_t$} \Comment{Update data loading proportion}\\
    }
    Sample a batch of data $\mathcal{B}$ from $S_{\text{train}, 1}, S_{\text{train},2}, \cdots, S_{\text{train},n}$ with proportion $w_t$\;
  Update $\theta$ with $\mathcal{L}(\theta, \mathcal{B})$ 
}
\sub{\update{$w$, $\Delta$}}{
  $\alpha \gets w \cdot \exp \left( \Delta \right)$ \Comment{Update  unnormalised weights} \\ 
  $w \gets \frac{\alpha}{\sum^n_i \alpha[i]}$ 
  \Return $w$ \Comment{Normalise domain weight}
}
\caption{\method{}}
\label{alg:domrescal}
\end{algorithm}
\end{figure}

\label{subsec:cpt}

\section{Experiment}
\label{sec:experiment}

In this section, we apply \method{} in continual pre-training CodeLlama-7B on the reasoning dataset and Llama3-8B as well as Mistral-7B on the system knowledge dataset.

\begin{figure*}[t]
\centering
\includegraphics[width=0.8\textwidth]{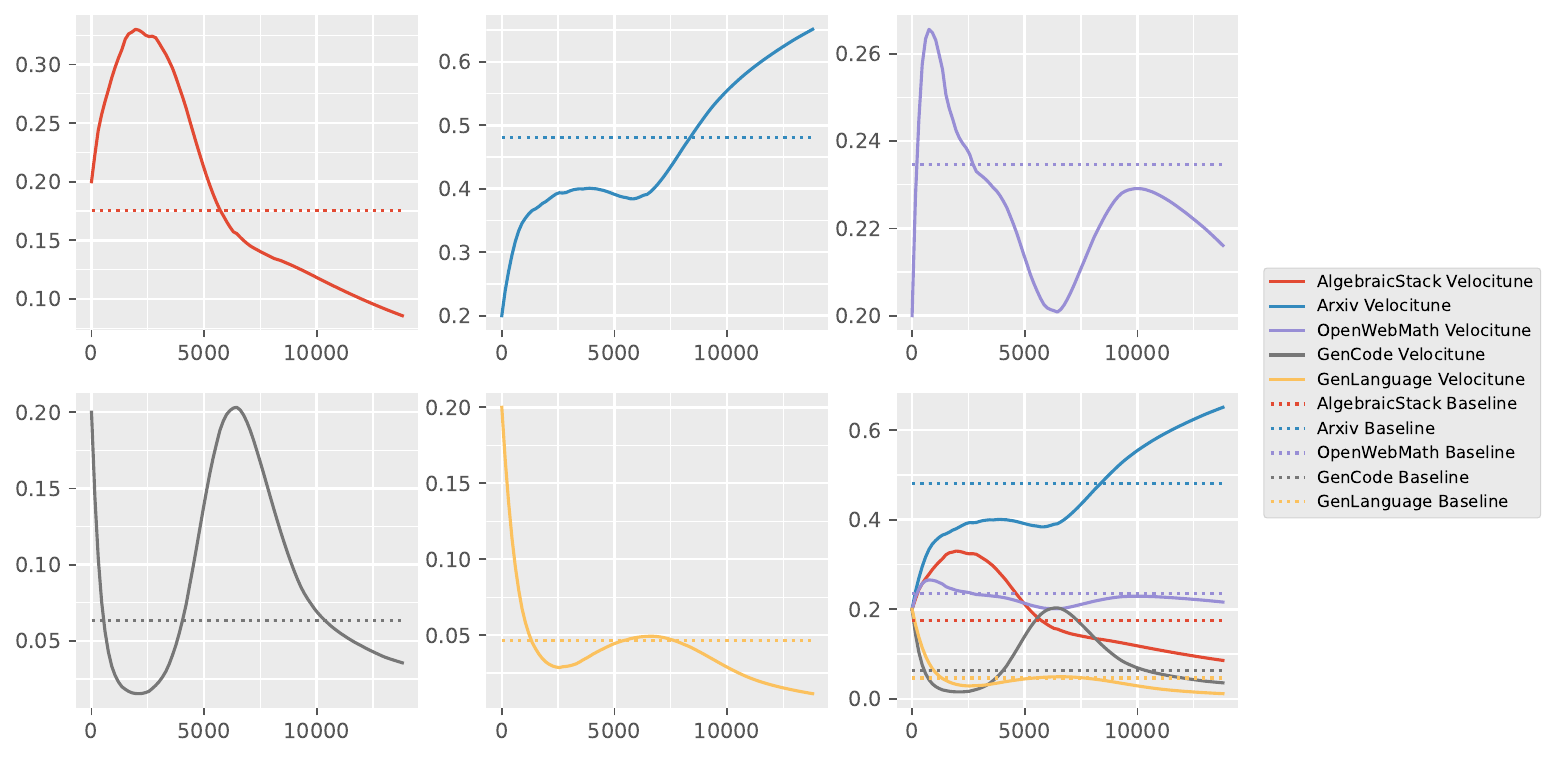}
\caption{Domain weights dynamic of \method{} in training CodeLlama 7B on \mathdata{}. The vertical axis represents domain weights, while the horizontal axis denotes training steps.}
\label{fig:mathcode}
\end{figure*}
\begin{table*}[!tb]
\caption{Performance of CodeLlama-7B, the baseline, and \method{} on multiple math and code benchmarks. The highest average accuracy for math and code benchmarks is highlighted in bold.}
\centering
\Large
\resizebox{0.8\textwidth}{!}{%

\begin{tabular}{lcccccc}

\toprule
\multirow{2}{*}{\textbf{Model}} & \multicolumn{6}{c}{\textbf{Math}}\\
\cmidrule{2-7} & \textbf{GSM8K}        & \textbf{MATH}         & \textbf{Minerva} & \textbf{MATHQA}    & \textbf{ASDiv}&\textbf{SVAMP}      \\
\midrule
CodeLlama& 12.4 & 6.0 & 5.20 & 14.1 & 50.5 &44.5\\
CodeLlama-Baseline & {28.9} & 11.1 & 9.80 & 24.0 & 61.1&{56.4} \\
CodeLlama-\method{} & 28.4 & {11.7} & {11.4} & {25.1} & 60.9&56.1 \\
\midrule \multirow{2}{*}{\textbf{Model}} &\multicolumn{3}{c}{\textbf{Math}}& \multicolumn{3}{c}{\textbf{Code}}\\
\cmidrule(lr){2-4} \cmidrule(lr){5-7} & \textbf{MMLU-STEM} & \textbf{SAT}     & \textbf{Math Avg.}& \textbf{HumanEval} & \textbf{MBPP} & \textbf{Code Avg.} \\\midrule
CodeLlama & 20.9 &18.8&21.6  &30.5 & 43.2 & 36.8 \\
CodeLlama-Baseline & 36.0 &46.9& 34.3 & 26.2 & 44.8& 35.5 \\
CodeLlama-\method{} & {37.3} &{56.2}& \textbf{35.9} \textcolor{red}{(+1.6\%)}  &{34.1} & {44.4} & \textbf{39.3} \textcolor{red}{(+3.8\%)} \\
\bottomrule
\end{tabular}
}

\label{tab:mathcode_res}
\end{table*}

\subsection{Experimental setup}

\paragraph{Training Corpus} We compile the \mathdata{} dataset based on Proof-Pile-2~\citep{azerbayev2023llemma} which consists of math reasoning text in natural language, format language, and code. The dataset includes three domains: {Arxiv}, {AlgebraicStack}, and {OpenWebMath}~\citep{Paster2023OpenWebMathAO}. Following the common practice of adding replay data to prevent catastrophic forgetting, we add two more domains: general code and general language which are composed of the Github subset from SlimPajama~\citep{cerebras2023slimpajama} and a blend of Slimpajama except Github and Arxiv. This results in a training set spanning five domains, with 76\% of the data in natural language and 24\% in code. For system knowledge, we use a dataset \textbf{SystemStack} which is collected from three domains {Arxiv}, {Blogs}, and {Stackoverflow}, concentrating on computer system-related knowledge. 
The statistics of the two datasets are shown in Table \ref{tab:dataset}.

\paragraph{Training Setup} We trained the models using the Negative Log-Likelihood (NLL) loss. Three settings were compared, \method{},  Dynamic Batch Loading (DBL)~\citep{xia2023sheared} which updates the weight using the distance between the evaluation loss and the target loss, and a baseline where sampling weights are determined by proportional token counts across domains during the continual pre-training. For DBL, we also apply the predicted target loss.
The total number of tokens processed during training was equivalent for the three methods to completing one full epoch using the training dataset. Detailed hyperparameters for the training process are summarised in \cref{appd:trainDetail} \cref{tab:training_details}.

\paragraph{Evaluation}To evaluate the math reasoning ability of models, we use $\text{math-lm-eval}$\footnote{https://github.com/ZubinGou/math-evaluation-harness} from ToRA~\citep{Gou2023ToRAAT} to evaluate the accuracy on GSM8K~\citep{Cobbe2021TrainingVT}, MATH~\citep{Hendrycks2021MeasuringMP}, Minerva~\citep{Lewkowycz2022SolvingQR},MMLU-STEM~\citep{Hendrycks2020MeasuringMM}, ASDiv~\citep{miao-etal-2020-diverse}, SVAMP~\citep{patel-etal-2021-nlp}, and MathQA~\citep{Amini2019MathQATI}. For coding ability, we use the evaluation kit from DeepSeek-Coder~\citep{Guo2024DeepSeekCoderWT} to assess the pass@1 accuracy on HumanEval~\citep{Chen2021EvaluatingLL} and MBPP~\citep{Austin2021ProgramSW}. 

For evaluating command generation ability, we use two benchmarks CmdGen-NVIDIA and CmdGen-AMD from the CmdGen series~\cite{lin2025sigmadifferentialrescalingquery}, which are built to assess the ability of models to provide proper system command when asked a related question. The two benchmarks, which provided a combination of 1.5K instruction-tuning data and 205 and 192 test questions respectively, evaluate model output from six metrics: 
Similarity of Command(\textbf{CMD Sim}): Cosine similarity of embeddings of generated and target commands. Similarity of Execution Output(\textbf{Output Sim}): Cosine similarity of embeddings of system outputs. Approximate Accuracy(\textbf{Approx Acc}): A value of 1 is assigned if either CMD Sim or Output Sim exceeds 0.9; otherwise 0, the overall Approx Acc is aggregated over test cases. The embeddings are generated by \textit{all-MiniLM-L6-v2}\footnote{https://huggingface.co/sentence-transformers/all-MiniLM-L6-v2}.
Exact Match (\textbf{EM}): The percentage of generated commands that are identical to the target commands.
Success Ratio (\textbf{SR}): The percentage of generated commands that produce the same system output as the target command.
Accuracy (\textbf{Acc}): The union of EM and SR, serving as the primary metric for the CMDGen task. Examples of CMDGen question and answer pair are shown in Appendix \ref{apdx:cmdgen}. 

\subsection{\mathdata{} training results}

\paragraph{\method{} learns math and maintains coding ability.}
In Table \ref{tab:mathcode_res}, we list the benchmarks tested to compare the methods in CodeLlama-7B with \mathdata{}. \method{} leads the baseline by 1.6\% on average across eight math benchmarks and 3.8\% on code average while the baseline's coding performance dropped by 1.3\% from before the continued pre-training. 
It underscores the reason behind \method{}'s effectiveness, which is balancing the learning velocity in each domain, while in the baseline due to the static domain mixture, the model might learn some domains well while not saturating for other domains. \method{} aligns the learning velocity across domains, resulting in more balanced learning progress, thus developing models of better downstream performance.

\begin{figure*}[t]
\centering
\includegraphics[width=0.75\textwidth]{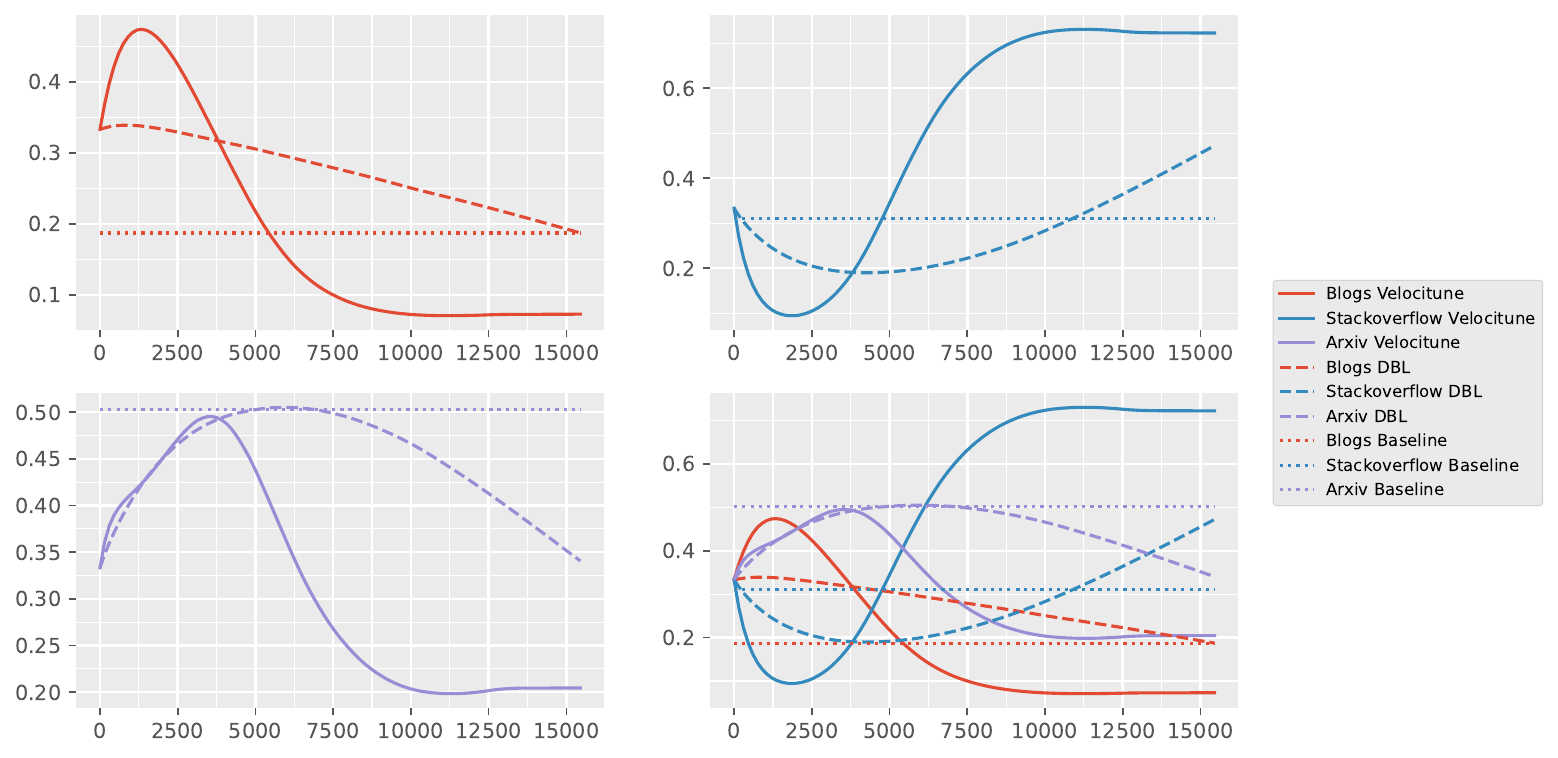}
\caption{Domain weights dynamic of \method{} and DBL in training Llama-3 8B on SystemStack. The vertical axis represents domain weights, while the horizontal axis denotes training steps.}
\label{fig:cmd}
\end{figure*}
\begin{table*}[!tb]
\small
\centering
\small
\caption{Results of \method{}, DBL, and Basline on Llama3 and Mistral on CmdGen-NVIDIA and CmdGen-AMD benchmarks. Best results for each metric are highlighted in bold.}
\resizebox{0.8\textwidth}{!}{%
\begin{tabular}{lcccccc}
\toprule
\multirow{2}{*}{\textbf{Model}}&    \multicolumn{6}{c}{\textbf{CmdGen-NVIDIA}}\\ 
\cmidrule(r){2-7} &   \textbf{Cmd Sim}&    \textbf{Output Sim}& \textbf{Approx Acc}& \textbf{EM}& \textbf{SR}& \textbf{Acc}\\
\midrule 
Mistral & 80.57&	58.67&	61.95&	24.88&	19.02&	30.73\\
Mistral-Baseline&  83.36&	65.26&	66.34	&23.90	&21.46	&32.20\\
Mistral-DBL&    79.83&	65.97&	59.02&	24.39&	19.02&	32.20 \\
Mistral-\method{}& 82.85&	64.30&	68.29&	27.32&	21.95&	36.59 \textcolor{red}{(+4.4\%)}\\
\midrule Llama-3 & 86.41&	69.09&	64.39&	41.95&	32.68&	50.73\\
Llama-3-Baseline&87.53&	72.15&	69.27&	46.34&	37.07&	57.07\\
Llama-3-DBL& 83.83&	64.87&	63.90&	38.54&	27.80&	45.85\\
Llama-3-\method{}&
\textbf{89.37}&	\textbf{75.59}&	\textbf{76.59}&	\textbf{51.21}&	\textbf{39.02}&	\textbf{61.95} \textcolor{red}{(+4.9\%)}\\
\midrule
\multirow{2}{*}{\textbf{Model}}&    \multicolumn{6}{c}{\textbf{CmdGen-AMD}}\\ 
\cmidrule(r){2-7} &   \textbf{Cmd Sim}&    \textbf{Output Sim}& \textbf{Approx Acc}& \textbf{EM}& \textbf{SR}& \textbf{Acc}\\
\midrule Mistral &84.25	&59.49&	61.54&	25.64&	15.90&	29.23 \\
Mistral-Baseline&   87.58&	65.72&	70.77&	25.13&	18.46&	30.77\\
Mistral-DBL&    84.61&	67.37&	65.13&	26.15	&17.95	&32.82 \\
Mistral-\method{}&  86.84	&62.99	&69.23	&27.18&	18.46&	33.33 \textcolor{red}{(+2.6\%)}\\
\midrule Llama-3&88.08&	71.22&	67.18&	41.54&	27.69&	47.18\\
Llama-3-Baseline&88.69&	69.60&	70.26&	46.15&	29.23&	51.79\\
Llama-3-DBL&    88.80&	68.00&	69.23&	39.49&	27.69&	45.64\\
Llama-3-\method{}&  \textbf{89.23}&	\textbf{72.97}&	\textbf{74.36}&	\textbf{49.74}&	\textbf{31.28}&	\textbf{54.87} \textcolor{red}{(+3.1\%)}\\
\bottomrule
\end{tabular}
}
\label{tab:cmd-res}
\end{table*}

\paragraph{\method{} dynamically regularises the learning velocity.} As shown in \cref{fig:mathcode}, the overall weight adjustments can be divided into three distinct stages. During the initial phase of training, the weights for newly introduced domains—OpenWebMath, Arxiv, and AlgebraicStack—rise, while those for the replay domains, GenCode and GenLanguage, decline sharply. This occurs because LMs are typically less saturated in the new domains, leading to slower learning compared to replay domains. As a result, \method{} increases the weights of the underperforming domains to balance their learning.
In the second stage, the weights for the replay domains begin to increase, while those of the new domains decrease. This shift occurs after the models have made substantial progress in the new domains, prompting \method{} to reallocate the focus to the replay domains.
In the final stage, the weights of the replay domains decrease again, while the downslopes for AlgebraicStack and OpenWebMath become more gradual, even showing a slight peak before continuing to decline. This suggests that the model has learnt these domains well. Meanwhile, \method{} increases the weight of Arxiv.
Adjusting weights dynamically, \method{} aligns the learning velocity across domains, ensuring a balanced learning outcome.


\subsection{SystemStack training results}
\begin{table*}[!h]
\caption{Performance comparison of CodeLlama-7B trained using \mathdata{} data with default domain ratios, reweighted domain ratios, and \method{}, evaluated on math and code reasoning benchmarks.}
\small
\centering
\resizebox{0.8\textwidth}{!}{%
\setlength{\tabcolsep}{3pt}
\begin{tabular}{lcccccc}
\toprule
\multirow{2}{*}{\textbf{Model}} & \multicolumn{6}{c}{\textbf{Math}}\\
\cmidrule{2-7} & \textbf{GSM8K}        & \textbf{MATH}         & \textbf{Minerva} & \textbf{MATHQA}    & \textbf{ASDiv       }&\textbf{SVAMP}      \\
\midrule
CodeLlama-Baseline & 28.9 & 11.1 & 9.8 & 24.0 & 61.1&56.4 \\
CodeLlama-Reweighted & 28.7 & 11.0 & 12.4 & 23.6 & 60.9 &56.3\\
CodeLlama-\method{} & 28.4 & 11.7 & 11.4 & 25.1 & 60.9&56.1 \\
\midrule \multirow{2}{*}{\textbf{Model}} &\multicolumn{3}{c}{\textbf{Math}}& \multicolumn{3}{c}{\textbf{Code}}\\
\cmidrule(lr){2-4} \cmidrule(lr){5-7} & \textbf{MMLU-STEM} & \textbf{SAT}     & \textbf{Math Avg.}& \textbf{HumanEval} & \textbf{MBPP} & \textbf{Code Avg.} \\\midrule
CodeLlama-Baseline & 36.0 &46.9& 34.3 & 26.2 & 44.8 & 35.5 \\
CodeLlama-Reweighted &37.7 &62.5& \textbf{36.6} & 30.5 & 42.8 & 36.6\\
CodeLlama-\method{} & 37.3 &56.2& 35.9   &34.1 & 44.4 & \textbf{39.3}\\ 
\bottomrule
\end{tabular}
}
\label{tab:mathcode_reweighted}
\end{table*}

\begin{table*}[!ht]

\centering
\small
\caption{Performance comparison of Llama3 models trained using SystemStack with original domain ratios, reweighted domain ratios, and \method{}, evaluated on the CmdGen benchmarks}
\resizebox{0.8\textwidth}{!}{%
\begin{tabular}{lcccccc}
\toprule
\multirow{2}{*}{\textbf{Model}}&    \multicolumn{6}{c}{\textbf{CmdGen-NVIDIA}}\\ 
\cmidrule(r){2-7} &   \textbf{Cmd Sim}&    \textbf{Output Sim}& \textbf{Approx Acc}& \textbf{EM}& \textbf{SR}& \textbf{Acc}\\ \midrule
Llama-3-Baseline&88.69&	69.60&	70.26&	46.15&	29.23&	51.79\\
Llama-3-Reweighted & 87.46&	72.09&	67.32&	46.83&	35.61&	56.10\\
Llama-3-\method{}&
\textbf{89.37}&	\textbf{75.59}&	\textbf{76.59}&	\textbf{51.21}&	\textbf{39.02}&	\textbf{61.95}\\
\bottomrule
\multirow{2}{*}{\textbf{Model}}&    \multicolumn{6}{c}{\textbf{CmdGen-AMD}}\\ 
\cmidrule(r){2-7} &   \textbf{Cmd Sim}&    \textbf{Output Sim}& \textbf{Approx Acc}& \textbf{EM}& \textbf{SR}& \textbf{Acc}\\ \midrule
Llama-3-Baseline&88.69&	69.60&	70.26&	46.15&	29.23&	51.79\\
Llama-3-Reweighted & 88.92&	\textbf{73.16}&	69.74&	47.18&	\textbf{32.31}&	\textbf{54.87}\\
Llama-3-\method{}&  \textbf{89.23}&	72.97&	\textbf{74.36}&	\textbf{49.74}&	31.28&	\textbf{54.87}\\
\bottomrule
\end{tabular}
}
\label{tab:cmd_reweighted}
\end{table*}

\paragraph{\method{} brings improvements across benchmarks.} On SystemStack, \method{} improves downstream task performance over DBL and Baseline on Llama3-8B and Mistral-7B.
Table \ref{tab:cmd-res} shows that \method{} improves the Accuracy in CMD-NVIDIA by 4.9\% and 4.4\% compared to baseline on Llama3-8B and Mistral-7B, and 3.1\% and 2.6\% and in CMDGen-AMD. We also found that Llama3-DBL underperforms the baseline on both benchmarks, which could be attributed to an imbalanced learning brought by updating domain weights based on the distances to the target loss.
\paragraph{\method{} accelerates weight stabilisation}
We plot the movement trajectory of the weights in SystemStack comparing \method{}, DBL, and the baseline in Figure \ref{fig:cmd}. In parallel with \citet{xia2023sheared}, we also observe that the weights stabilise after a few thousand steps. As re-scaling accelerates convergence in machine learning~\citep{juszczak2002feature}, we also observe that \method{} on SystemStack reaches stabilisation at least 1.5x faster than DBL as DBL does not stabilise at the end of training. 
We hypothesise that if the DBL training continues for more steps, it will reach the same final weights as \method{}. For training on \mathdata{} which is shown in \cref{fig:mathcode}, we do not observe a complete stabilisation, but only the curve becoming flat at the end of training.

\subsection{Data ordering contributes along with reweighted domain weights}

To isolate the key contributors behind the effectiveness of \method{}, we performed an ablation study to examine the effect of the reweighted data proportion. Specifically, we collect the weights during \method{} for each evaluation interval and then average the weights to get the overall sampling ratio of each domain during training. A comparison of the original ratio and the reweighted ratio is shown in Tables \ref{tab:system_reweight_ratio} and \ref{tab:pp2_reweight_ratio}. Using the reweighted data ratio, we train the model on the dataset using the static mixture for the same training steps. We conducted the experiment for CodeLlama-7B on the \mathdata{} dataset and Llama3 on SystemStack. The results are shown in Table \ref{tab:mathcode_reweighted} and Table \ref{tab:cmd_reweighted}.

\begin{table*}[!htb]
\caption{Performance comparison of CodeLlama-7B trained using \mathdata{} data with or without target loss and Baseline, evaluated on math and code benchmarks.}
\centering
\Large
\resizebox{0.8\textwidth}{!}{%
\begin{tabular}{lcccccc}
\toprule
\multirow{2}{*}{\textbf{Model}} & \multicolumn{6}{c}{\textbf{Math}}\\
\cmidrule{2-7} & \textbf{GSM8K}        & \textbf{MATH}         & \textbf{Minerva} & \textbf{MATHQA}    & \textbf{ASDiv       }&\textbf{SVAMP}      \\
\midrule
\method{} w/o target loss&17.5&	8.2&	7.4&	31.5&	43.8&	48.8\\
Baseline & 28.9 & 11.1 & 9.80 & 24.0 & 61.1&56.4 \\
\method{} & 28.4 & 11.7 & 11.4 & 25.1 & 60.9&56.1 \\
\midrule \multirow{2}{*}{\textbf{Model}} &\multicolumn{3}{c}{\textbf{Math}}& \multicolumn{3}{c}{\textbf{Code}}\\
\cmidrule(lr){2-4} \cmidrule(lr){5-7} & \textbf{MMLU-STEM} & \textbf{SAT}     & \textbf{Math Avg.}& \textbf{HumanEval} & \textbf{MBPP} & \textbf{Code Avg.} \\\midrule
\method{} w/o target loss &	54.2&	30.2& 30.2& 28.0 & 42.0& 35.0\\
Baseline & 36.0 &46.9& 34.3 & 26.2 & {44.8} & 35.5 \\
\method{} & 37.3 & {56.2}& \textbf{35.9}  &{34.1} & 44.4 & \textbf{39.3} \\
\bottomrule
\end{tabular}}
\label{tab:target_loss}
\end{table*}

On CmdGen benchmarks, Llama-3 trained with reweighted ratio data(denoted as Llama-3-Reweighted) outperforms Llama3-Baseline on both benchmarks, but is inferior to \method{} on NVIDIA and on par with \method{} on AMD.
In the evaluation of math and code reasoning, CodeLlama trained with the reweighted ratio(denoted as CodeLlama-Reweighted) achieves a slightly higher accuracy than \method{} on the math benchmarks, but still falls behind \method{} by 2.7\% for the coding benchmark.
In general, reweighted data ratios improve downstream task performance compared to the original mixture. However, the models still underperform relative to \method{}, with the sole exception being the Math average, where the performance is 0.7\% higher than \method{}. This finding sheds new light on the common assumption that data mixture is the primary factor driving downstream performance. 
Previous studies~\citep{Hu2024MiniCPMUT, llama3modelcard} have highlighted the effects of presenting different data at various stages of model training. Our comparison further supports the potential of dynamic data weighting during pre-training. We encourage further research into the impact of data ordering and its role in optimising model performance.


\begin{figure*}[!h]
\centering
\includegraphics[width=2\columnwidth]{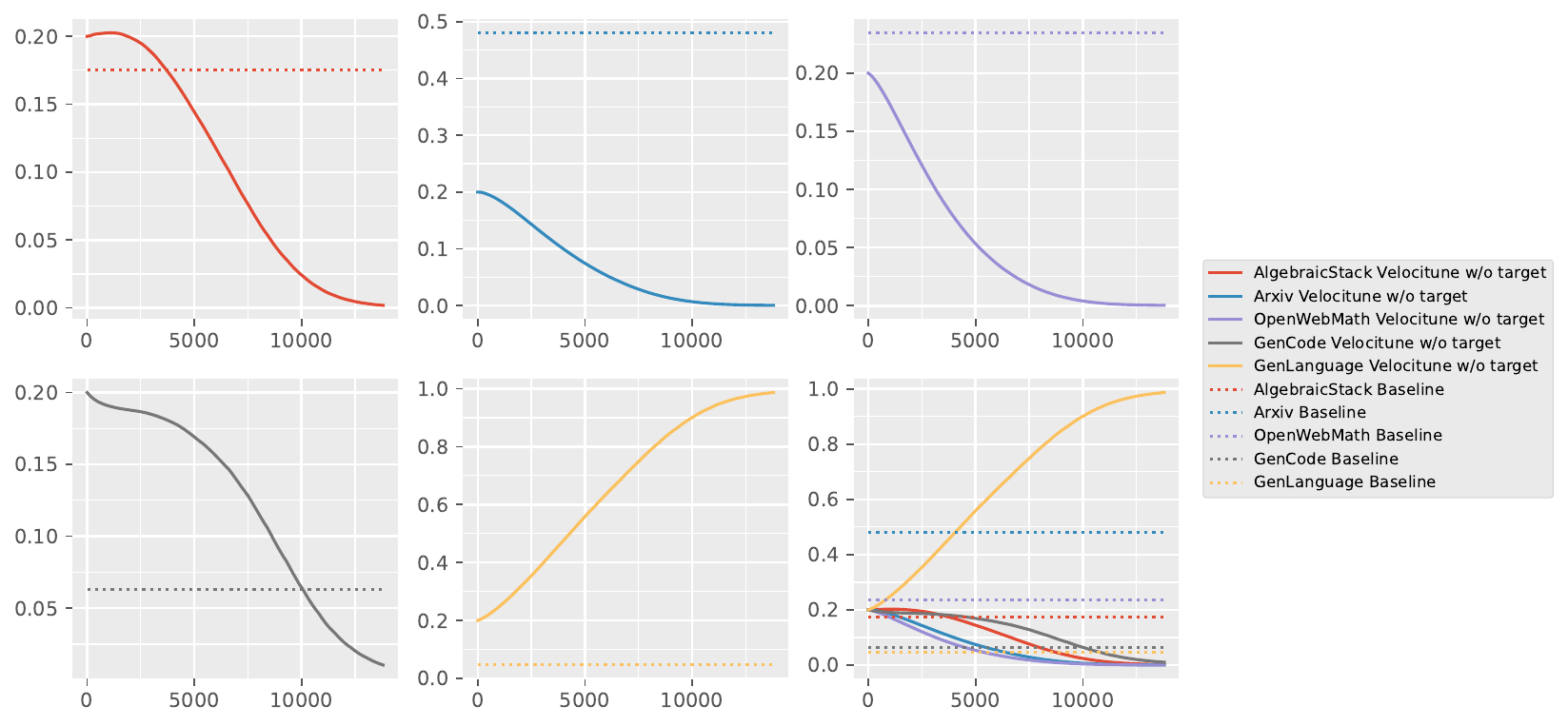}
\caption{Domain weights dynamic of \method{} without target loss in training CodeLlama 7B on Math\&Code.}
\label{fig:inverse}
\end{figure*}

\subsection{Target loss is critical to \method{}}
To highlight the significance of the target loss estimate in learning velocity, we also explore an alternative method that excludes the target loss in learning velocity when updating the weights. In this setting, Equation \ref{eq:velocity} is reduced to dividing the evaluation loss by the initial loss for each domain, as given by:
$\delta^i = \frac{\ell_t^i(\theta, S^{\prime}_{\text{eval}, i})}{\ell^i_{\text{init}}}$. Using this formulation, we train CodeLlama on the \mathdata{} dataset.
The evaluation results are presented in \cref{tab:target_loss} and the dynamic of the domain weight is shown in \cref{fig:inverse}. We see that without the constraints of target loss, the weight of GenLangauge is driven to nearly 1 and others are suppressed to nearly 0. Consequently, the model exhibits a performance decrease of 4.1\% in the math benchmarks and 0.5\% in the code benchmarks compared to the baseline. We hypothesise that this update function implicitly assumes a target loss of zero for each domain, leading to unrealistic expectations for the model to minimise the loss to zero across all domains. Such impractical targets disrupt the training process, ultimately degrading overall performance.



\section{Related Work}
\label{sec:related}
GLaM~\citep{du2022glam} manually assigns weights to each domain based on the amount of data and the downstream performance of a small model trained on the domain data. Lemma~\citep{azerbayev2023llemma}, an effort to continue pre-training LMs with math-related data from three domains, determines domain weights by iterating through different combinations and selecting those with the lowest perplexity on an evaluation dataset. Doremi~\citep{xie2024doremi} employs a three-stage process to dynamically adjust domain weights: (1) Train a reference model, (2) Train a proxy model while adjusting the proportion of data based on the proxy model's loss, and (3) Train the final model using the aggregated data weights, demonstrating superior performance compared to models trained with the original weights. 
\citet{xia2023sheared} shows that instead of training a reference model, the target loss value can be estimated using scaling laws~\citep{hoffmann2022training}, achieving better performance on smaller LMs with the predicted losses.


\section{Conclusion}
\label{sec:conclusion}
In this work, we introduced \method{}, a novel approach to dynamically adjust domain weights during the training of large language models. Our method addresses the gap of existing works in domain-adapative continual pre-training. By aligning learning velocity across domains, \method{} ensures a more balanced learning process, thereby enhancing the model's ability to generalise across diverse downstream tasks.
Our experiments on CodeLlama-7B, Llama3-8B, and Mistral-7B demonstrate the effectiveness of \method{} in improving performance in a variety of tasks, including math reasoning, coding, and system command generation.  Furthermore, our ablation study reveals that the benefits of \method{} are not solely due to the adjustment of the domain weights, but also stem from the synergistic effect of the training dynamics, including the order in which data from different domains are processed. This finding underscores the importance of considering both the weights and the data order of in continual pre-training.

\section{Limitations}
The limitations of our study are threefold. First, \method{} is currently designed for continual pre-training and has not yet been evaluated in pre-training from scratch. Second, while supervised fine-tuning (SFT) datasets often span multiple domains, the applicability of \method{} in the SFT stage remains unexplored. Third, Velocitune incurs higher costs than the baseline due to its evaluation process during training. However, since pre-training primarily prioritises downstream performance, we defer a detailed analysis of computational efficiency and potential optimisations of \method{} to future work.

\section{Ethics Statement}
This research is based on open-source models and is intended solely for research purposes. We acknowledge potential risks.
While Velocitune improves model efficiency and performance, improved reasoning and code generation capabilities could be misused in misinformation generation, automated decision making, or bias reinforcement.
Our work focusses on specific domains (maths, code, and system commands), which may not generalise well to under-represented languages or topics.
Our research does not involve sensitive or private data.



\bibliography{custom}

\newpage
\appendix

\section{Examples of CmdGen}
We show in the following two question and answer pairs from the CMDGen benchmarks.
\label{apdx:cmdgen}
\begin{tcolorbox}
\footnotesize
\textbf{CMDGen-NVIDIA}

\textbf{Question}: \textcolor{blue}{How to compile a CUDA file with a restriction on register count per thread to 32 and enable verbose output?}

\textbf{Target Command}: \textcolor{blue}{nvcc -v -maxrregcount=32 sample.cu}

\end{tcolorbox}

\begin{tcolorbox}
\footnotesize
\textbf{CMDGen-AMD}

\textbf{Question}: \textcolor{blue}{How to use 'rocm-smi' to show the range of memory clock frequencies (mclk) that can be set?}

\textbf{Target Command}: \textcolor{blue}{rocm-smi --showmclkrange}

\end{tcolorbox}


 

      


\section{Scaling law prediction details}

We continue training until the difference between the predicted target losses using the previous checkpoint sets and the current checkpoint set becomes smaller than a specified threshold (denoted $\sigma$ ). This criterion helps to select an appropriate value for k during the proxy model training.
Detailed scaling law prediction errors are given in \cref{tab:scalaw}.
\label{appd: scalaw}
\begin{table}[!h]
\small
\centering
\caption{The average error over domain for predicting evaluation loss of models using full training data. We used evaluation loss from checkpoints saved till using half the training data to fit the laws.}
\setlength{\tabcolsep}{6pt}
\begin{tabular}{lcc}
\toprule
&\textbf{\mathdata{}}&\textbf{ SystemStack}\\
\midrule
Error&2.4e-3&1.84e-3\\
\bottomrule
\end{tabular}
\label{tab:scalaw}
\end{table}

\section{Training datasets statistics}
The statistics of the training corpus are shown in \cref{tab:dataset}.

        

    

\begin{table}[!h]
    \caption{Statistics for Proof-Pile-2Plus and SystemStack. Token counts were determined using the CodeLlama and Llama3 tokenizers, respectively.}
    \label{tab:dataset}

      \centering
    \begin{tabular}{cc}
    \toprule
    \multicolumn{2}{c}{\textbf{\mathdata{}}}\\
    \midrule \textbf{Domain} & \textbf{\# Tokens} (Billion) \\
         \midrule Arxiv & 28.70\\
         AlgebraicStack & 10.47\\
         OpenWebMath & 14.02\\
         GenCode & 3.97\\
         GenLangauge& 2.92\\
    \bottomrule
    \end{tabular}

      \centering
    \begin{tabular}{cc}
    \toprule
        \multicolumn{2}{c}{\textbf{SystemStack}}\\
    \midrule \textbf{Domain} & \textbf{\# Tokens} (Billion) \\
      \midrule Arxiv & 5.37 \\
     Blogs & 3.21\\
     StackOverflow & 7.64\\
     \bottomrule
    \end{tabular}

\end{table}

\section{Training Details}
\label{appd:trainDetail}
We run the experiments on 64 Nvidia H100 GPUs. Our distributed training is based on \textit{accelerate}\footnote{\url{https://huggingface.co/docs/accelerate/en/index}} and \textit{FDSP}\footnote{\url{https://pytorch.org/blog/introducing-pytorch-fully-sharded-data-parallel-api/}}.
\begin{table}[!h]
    \centering
    \small
    \caption{Training hyper-parameters and throughput.}
    \begin{tabular}{lcc} \toprule
    &SystemStack&\mathdata{}\\
   \midrule Training tokens                                                 & $16$B     & $63$B                   \\
    Learning rate                          & $1e-5$ & $5e-5$              \\
    LR warmup ratio                                                 & $0.005$     & $0.005$                   \\
    Batch size        (tokens)                                              & $1$M     & $4$M                    \\
    Evaluation interval $m$       (steps)                                 & $150$       & $150$               \\ 
    Steps & $15,482$ & $13,807$ \\
    \# GPUs & $64$ & $64$ \\
    Sample ratio for proxy model& 58\% & 51\%\\
    Adam $\beta_2$ &0.95&0.99\\
    \bottomrule 
    \label{tab:training_details}
    \end{tabular}
\end{table}

\section{Reweighted Data Ratio after \method{}}
The reweighted data ratio is shown in \cref{tab:system_reweight_ratio} and
\cref{tab:pp2_reweight_ratio} 
\begin{table}[!h]
    \centering
    \caption{The domain weights of Baseline and \method{}'s reweighted domain weights on SystemStack for Llama3}

    \begin{tabular}{lcccc}
    \toprule
    & Blogs& Stackoverflow& Arxiv\\
    \midrule
    Baseline & 0.187 & 0.503 & 0.310\\
    Reweighted & 0.197 & 0.470& 0.333 \\
    \bottomrule
    \end{tabular}
    \label{tab:system_reweight_ratio}
\end{table}

\begin{table}[!h]
\small
    \caption{The domain weights of Baseline and \method{}'s reweighted domain weights on \mathdata{} for CodeLlama}
    \centering
    \resizebox{\columnwidth}{!}{%
    \begin{tabular}{lcccccc}
    \toprule
    & AlgebraicStack& Arxiv& OpenWebMath&GenCode&GenLanguage\\
    \midrule
    Baseline & 0.189 & 0.500 & 0.259&0.029&0.020\\
    Velocitune & 0.185 & 0.463 & 0.225&0.086&0.040\\
    \bottomrule
    \end{tabular}}
    \label{tab:pp2_reweight_ratio}
\end{table}
.


\end{document}